\documentclass{article}
\usepackage{ijcai26}

\usepackage{times}
\usepackage{soul}
\usepackage{url}
\usepackage[hidelinks]{hyperref}
\usepackage[utf8]{inputenc}
\usepackage[small]{caption}
\usepackage{amsmath}
\usepackage{amsthm}
\usepackage{amsfonts}
\usepackage{amssymb} % 提供 \checkmark

\usepackage{graphicx}
\graphicspath{{figures/}}

\usepackage{subcaption} 

\usepackage{booktabs} 
\usepackage[table]{xcolor} % 放在这里处理表格颜色，支持 \rowcolor

\usepackage{pifont} % 提供 \ding{55} 叉号
\usepackage[switch]{lineno}

\usepackage[ruled,linesnumbered,algo2e]{algorithm2e}

\SetAlgoNoLine               % 关键：去掉那些令人不忍直视的垂直杠
\SetNlSty{textrm}{}{:}       % 行号改为 1: 样式，使用正文字体
\SetKwSty{textbf}            % 关键字（while, for）仅加粗，不使用斜体
\SetAlgoNlRelativeSize{-1}   % 行号稍微缩小，更显精致
\SetNlSkip{0.5em}            % 行号与代码的间隙
\SetInd{0.5em}{1em}          % 严格缩进控制，修正了 \SetInd 的大小写报错
\SetArgSty{textrm}           % 循环条件（如 not converged）使用正文样式而非斜体 \Setind -> \SetInd
\usepackage{multirow}      % 必须加载，用于合并多行

\title{CASE-NET: Deep Spatio-Temporal Representation Learning via Causal Attention and Channel Recalibration for Multivariate Time Series Classification}

\author{
	Fan Zhang$^1$
	\and
	Yating Cui$^1$
	\and
	Hua Wang$^2$\\
	\affiliations
	$^1$Shandong Technology and Business University\\
	$^2$Ludong University\\
	\emails
	zhangfan51@sina.com,
	20254020269@sdtbu.edu.cn,
	hwa229@163.com
}

\begin{document}

\maketitle

\begin{abstract}
	Multivariate time series (MTS) classification is foundational to pervasive computing and financial analysis, yet existing multi-scale paradigms are often constrained by suboptimal representation fidelity. We identify two critical bottlenecks: temporal non-causality in standard encoders that induces temporal confounding in non-stationary dynamics, and the absence of explicit channel saliency mechanisms that allows noise to contaminate the latent space. To address these challenges, we propose the \textbf{C}ausal \textbf{A}ttention and \textbf{S}patio-temporal \textbf{E}ncoder \textbf{Net}work (\textsc{CASE-NET}), an architecture designed for structural manifold pre-conditioning. \textsc{CASE-NET} synergizes a Causal Temporal Encoder, which enforces physical arrow-of-time constraints via masked self-attention and causal convolutions, with an Adaptive Channel Recalibration module functioning as an information bottleneck to suppress detrimental noise. Comprehensive evaluations across six heterogeneous domains demonstrate that \textsc{CASE-NET} establishes new state-of-the-art benchmarks on four tasks, achieving a peak accuracy of 98.6\% on the AWR dataset and superior robustness in non-stationary regimes.
\end{abstract}
\section{Introduction}

The surge in IoT and wearable data makes MTS classification a core AI challenge, ranging from clinical diagnostics to activity recognition~\cite{morid2023time,liu2024spatial}. Unlike static data modalities, MTS exhibits profound Spatio-Temporal Complexity: temporal dimensions encapsulate non-stationary dynamics across multiple granularities, while spatial dimensions involve intricate, high-dimensional couplings between heterogeneous variables~\cite{wang2024deep}. Extracting robust representations from these complex sequences is a core challenge.

Multi-scale analysis has proven effective in capturing cross-granularity patterns. Building upon this, recent breakthroughs in Multi-scale Disentanglement have successfully introduced scale-shared and scale-specific orthogonal constraints to eliminate inter-scale redundancy~\cite{liu2025disms}, fostering a solid theoretical foundation for representation learning. However, these frameworks typically operate under the implicit assumption that the spatio-temporal features fed into the disentanglement module are sufficiently robust and noise-resilient. Critically, we argue that this assumption is often violated in real-world scenarios due to two representation bottlenecks in current feature extraction pipelines. First, existing encoders often suffer from temporal blindness: recurrent models struggle with long-range dependencies, while bidirectional processing may introduce future-induced stochastic fluctuations into current representations. In physical systems, the arrow of time dictates that current states should not depend on future observations. For non-stationary MTS, such future observations often correspond to evolutionary noise or sensor artifacts rather than useful semantic context, leading to temporal confounding and distorted latent motifs.

Second, high-dimensional sensing systems also exhibit spatial channel heterogeneity. Variable importance is inherently dynamic and context-dependent, yet many architectures treat all channels with homogeneous priority and leave noise filtering to the downstream disentanglement module. This increases the burden on the disentanglement head and fundamentally limits the purity of the distilled features. CASE-NET addresses these two bottlenecks by redefining the causal mask as a structural noise filter and introducing channel recalibration as a proactive spatial information bottleneck.

Beyond time-series analysis, compositional representation learning has also been actively explored in composed image and video retrieval, where models disentangle target semantics from modification cues to support robust cross-modal matching~\cite{ENCODER,PAIR,MEDIAN,Chen2025hud,zhang2026hint,ReTrack,ConeSep}. Inspired by this general principle of composition-aware representation learning, we investigate compositionality along temporal-scale and channel dimensions in MTS, where future-induced stochasticity and sensor heterogeneity become the dominant obstacles. Motivated by the insight that better representation facilitates better disentanglement, we propose CASE-NET. Our objective is to implement structural manifold pre-conditioning on the upstream representations to ensure high-fidelity feature decoupling. Our contributions are summarized as follows:

\begin{itemize}
	\item {Causal-Aware Temporal Reconstruction:} We introduce a causal attention framework that explicitly enforces a physical arrow-of-time prior, filtering future-induced evolutionary noise to ensure the integrity of latent {action motifs}. By coordinating local causal convolutions with global masked self-attention, the CASE encoder functions as a structural {manifold pre-conditioner} that is architecture-agnostic, empowering various downstream tasks as a high-fidelity front-end.
	
	\item {Context-Adaptive Spatial Recalibration:} We implement a channel-wise {information bottleneck} via adaptive recalibration (SE module) to execute proactive {manifold pre-conditioning}. By purging non-discriminative spatial artifacts before the disentanglement stage, this mechanism ensures that only high-saliency variables propagate to the latent space, thereby enhancing the purity of the distilled features.
	
	\item {Holistic Evaluation across Six Heterogeneous Domains:} Unlike conventional approaches validated only in a single domain, \textsc{CASE-NET} was fully evaluated on six benchmark datasets with diverse physical characteristics and data distributions. Results show that \textsc{CASE-NET} established new state-of-the-art benchmarks on four tasks.
\end{itemize}

\section{Related Work}

\subsection{Deep Temporal Modeling}
Extracting temporal dependencies is central to MTS classification~\cite{qin2017dual,zhu2023long}. While traditional recursive models like LSTMs and GRUs utilize inductive bias for sequential encoding, they often suffer from vanishing gradients and a critical \textit{paradigm limitation}: {Temporal Blindness}. Most recurrent architectures lack strict physical causality, leading to spurious future-information leakage that compromises reliability in sensitive domains like clinical diagnostics. Although recent works attempt to encode causality through transfer entropy \cite{sun2024deep} or spatiotemporal information transformation \cite{cai2025causal}, these often incur heavy computational costs. \textsc{CASE-NET} addresses this by re-engineering the encoder with causal convolutions and masked self-attention, enforcing strict temporal arrows to capture global dynamic motifs with high fidelity. Recent forecasting models further exploit multi-offset temporal interactions, semantic fusion, and non-ideal data utilization to strengthen multivariate dependency learning~\cite{zhang2026time,zhang2026timesaf,qiu2025duet,wang2026idealtsf}. Nevertheless, these methods primarily emphasize predictive dependency modeling, whereas CASE-NET focuses on classification-oriented representation purification under strict temporal causality. In addition, ensuring high-fidelity representations remains a challenge, as explored by Zhang et al.~\cite{zhang2022self} through time-frequency consistency. In contrast, \textsc{CASE-NET} enforces a strict causal inductive bias to purify manifolds, especially in nonstationary states. This distinguishes it from latent SDE-Attention~\cite{fang2025sde}, which focuses on irregular sampling, as we optimize the upstream representation manifold for deterministic MTS. Recently, Xu et al.~\cite{xu2025deep} proposed a spatio-temporal fusion architecture for dynamic ECN analysis, using Granger-causality-based discovery to model the evolution of brain networks. While their approach focuses on learning the explicit causal structure from observational data, \textsc{CASE-NET} leverages a \textit{causal inductive bias} through architectural masking. We argue that for non-stationary classification tasks, enforcing causality as a structural prior is more effective for filtering future-induced stochastic noise than explicit causal discovery, which can be sensitive to latent confounding factors.

\subsection{Channel Saliency and Information Bottleneck}
Attention mechanisms have shifted MTS modeling toward dynamic recalibration. Recent architectures have introduced cross-channel attention \cite{hao2020new,zhang2025multivariate} or association adapters \cite{lin2025cats} to handle multivariate dependencies. Recent developments such as iTransformer \cite{liu2024itransformer} have rethought the role of channel dependencies by inverting the transformer structure. However, our approach differs in the implementation of active information bottlenecks through adaptive recalibration to actively suppress noise rather than focusing solely on channel-independent projections. However, many frameworks still rely on a suboptimal assumption of \textit{Spatial Channel Homogeneity}, treating sensors with uniform priority regardless of their varying Signal-to-Noise Ratios~\cite{cheng2023weakly,huang2023crossgnn}. Consistent with feature-centric explanation models like CAFO \cite{kim2024cafo}, \textsc{CASE-NET} leverages {Information Bottleneck} theory and integrates an adaptive recalibration mechanism to execute proactive feature-cleaning, ensuring only high-saliency variables propagate to the latent space for robust disentanglement.

\subsection{Multi-scale Analysis and Disentanglement Fidelity}
Recent milestones like DisMS-TS have introduced multi-scale disentanglement to mitigate inter-scale redundancy~\cite{liu2025disms}. At the benchmark and optimization levels, recent studies have emphasized fair time-series evaluation, non-ideal data utilization, and scalable optimization for time-series modeling~\cite{qiu2024tfb,wang2026idealtsf,wang2026eeo,qiu2025DBLoss}. Unlike multi-scale token mixing \cite{zhong2025mtm}, hierarchical temporal-frequency fusion \cite{song2025multi}, or multi-view meta-learning \cite{zhang2024multiview} that prioritize feature fusion, the efficacy of decoupling is fundamentally contingent upon the \textit{fidelity of underlying representations}. Existing frameworks often overlook feature extractor quality,         creating a performance bottleneck when processing noise-contaminated or temporally inconsistent inputs. Departing from efforts focused solely on downstream decoupling logic, \textsc{CASE-NET} focuses on optimizing the quality of the upstream representation manifold through structural pre-conditioning. By providing causal-aware spatio-temporal priors, our model bridges the gap between raw sensing data and decoupled representations,           establishing new performance ceilings across heterogeneous domains.

\section{Methodology}

\subsection{Overall Architecture}
CASE-NET enhances classification through three stages (Fig. 1): Multi-scale Projection, the CASE Encoder, and decomposition. The CASE encoder is weight-shared across all $S$ branches. This ensures that the model learns a scale-invariant causal-spatial transformation while maintaining efficiency.

{Formal Rationale:} We define the CASE encoder not merely as a collection of components, but as a \textit{synergistic information bottleneck} and a \textit{manifold pre-conditioner} designed to address the fundamental problem of \emph{distorted input manifolds} in existing frameworks. Existing multi-scale encoders often suffer from {temporal blindness} and {channel homogeneity}, which introduce future-information leakage and spatial noise. Such contaminated inputs force the downstream disentanglement loss ($\mathcal{L}_{diff}$) to search within an excessively complex hypothesis space. By purifying the input manifold through structural pre-conditioning before it enters the sensitive disentanglement backend, we significantly narrow the search space of the downstream loss function and ensure high-resolution representations.

\begin{figure*}[t]
	\centering
	\includegraphics[width=0.9\linewidth]{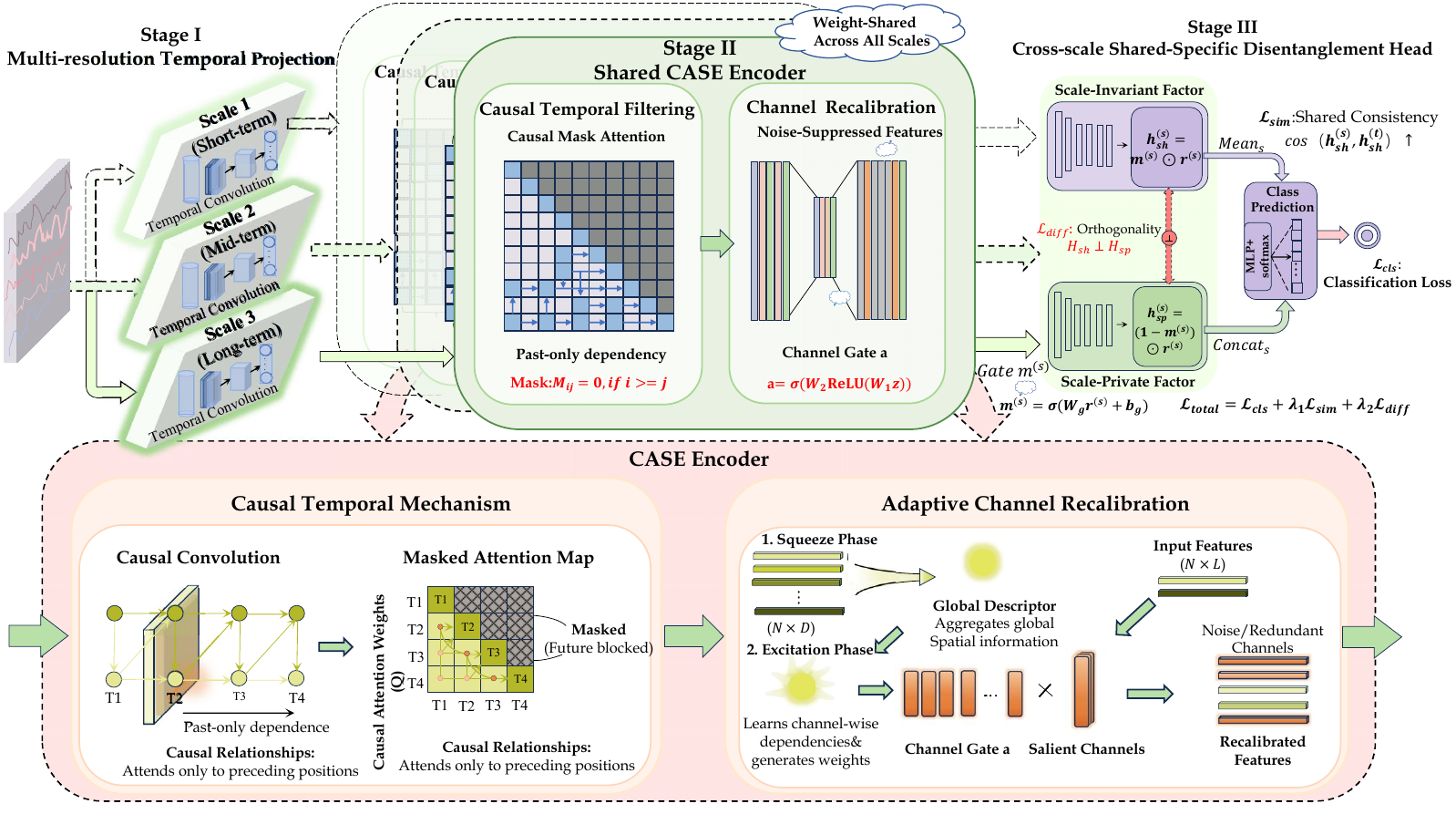} 
	\caption{The hierarchical framework of \textsc{CASE-NET}, illustrating: (1) multi-scale representation initialization via parallel branches; (2) a \textit{weight-shared} CASE encoder (2 layers) that processes all branches synchronously, constrained by physical laws and SNR; and (3) inter-scale redundancy elimination via orthogonal constraints.}
	\label{fig:overall_architecture}
\end{figure*}

\subsection{Generic Multi-scale Projection}
To capture the polymorphic feature mapping of MTS at different granularities, we construct a multiscale input space consisting of $S$ branches. Specifically, given the input $\mathbf{X} \in \mathbb{R}^{N \times L}$, we utilize a set of convolutional operators with heterogeneous receptive fields and strides to project it as a multi-resolution view set $\mathcal{V} = \{\mathbf{V}_s\}_{s=1}^S$, where:
\begin{equation}
	\mathbf{V}_s = \sigma ( \text{Conv1D}_s(\mathbf{X}) ) \in \mathbb{R}^{C_s \times L_s}
\end{equation}
Non-uniform sampling here establishes multi-granularity representations, providing context for the CASE encoder. Importantly, this stage only initializes scale-specific views, while the subsequent weight-shared CASE encoder imposes a unified causal-spatial transformation across all resolutions.

\subsection{Causal Adaptive Spatio-temporal Encoder}
The CASE encoder serves as the core architecture for spatio-temporal feature extraction, specifically designed to solve the \textit{representation fidelity bottleneck} in multi-scale decoupling. Rather than a standard architectural stack, the encoder functions as a structural pre-conditioner that integrates two synergistic modules: Causal Temporal Attention and Channel Adaptive Recalibration. This dual-constrained design ensures that the resulting latent space is both physically consistent and semantically sparse.

{3.3.1 Causal Temporal Attention.}
The CTA module is designed to overcome the bottlenecks of traditional recurrent models in long-range dependency modeling while imposing a robust temporal causality constraint. By coupling local causal operators (Dilated Convolutions) with global masked self-attention, it ensures autoregressive consistency ($\mathbf{h}_t = f(\{x_\tau\}_{\tau \le t})$). This configuration eliminates future information leakage while capturing multi-scale, long-range dependencies across the sequence.

In contrast to bidirectional transformers or explicit discovery models \cite{xu2025deep}, \textsc{CASE-NET} recognizes that in non-stationary regimes (e.g., NASDAQ), future observations often represent stochastic evolutionary noise relative to the current state. In contrast to bidirectional transformers (e.g., BERT) where future context helps clarify semantics, in high-frequency non-stationary MTS, future observations often contain stochastic noise that does not reflect core discriminative motifs. By enforcing causality, our CTA module functions as a {Structural Noise Filter}. This ensures the encoder prioritizes the capture of underlying \textit{action motifs} rather than fitting transient \textit{signal fluctuations}, effectively executing {Temporal De-confounding} through architectural constraints. By enforcing this {Manifold Pre-conditioning}, we narrow the downstream hypothesis space and ensure the representation manifold remains invariant to future-induced noise, preventing any form of data leakage within the architecture.

To instantiate local causal modeling, we introduce dilated convolutional layers with asymmetric left padding. For an input sequence $\mathbf{X}$, the temporal dimension is padded with $L_{pad}=(k-1)\times d$ zeros on the left, where $k$ denotes the kernel size and $d$ is the dilation rate. The causal operator is defined as:
\begin{equation}
	\mathbf{y}_t = \sigma \left( \sum_{i=0}^{k-1} \mathbf{W}_i \cdot \mathbf{x}_{t-d \cdot i} + \mathbf{b} \right)
\end{equation}
This operator forces the convolution kernel to slide only over current and historical trajectories, extracting local dynamics with high causal fidelity. Such a structured inductive bias is crucial for non-stationary sequences, since it keeps the transition probability $P(x_t | x_{<t})$ undisturbed by future observations.

To complement local causal filtering, we further integrate multi-head self-attention with an upper-triangular causal mask $\mathbf{M}\in\mathbb{R}^{L\times L}$. The mask is defined as:
\begin{equation}
	M_{ij} = \begin{cases} 0, & i \ge j \\ -\infty, & i < j \end{cases}
\end{equation}
Given linear projection matrices $\mathbf{W}^Q$, $\mathbf{W}^K$, and $\mathbf{W}^V$, the masked attention reconstructs the temporal manifold by:
\begin{equation}
	\text{Attention}(\mathbf{Q}, \mathbf{K}, \mathbf{V}) = \text{Softmax}\left( \frac{\mathbf{Q}\mathbf{K}^\top}{\sqrt{d_k}} + \mathbf{M} \right) \mathbf{V}
\end{equation}

{3.3.2 Channel Adaptive Recalibration.}
To complement the temporal consistency established by CTA, this module instantiates the {Information Bottleneck} principle to address {Channel Heterogeneity}. Due to varying signal-to-noise ratios (SNRs) across sensors, the raw temporal representation $\mathbf{H}$ is frequently contaminated by non-discriminative channel noise.

The recalibration module complements CTA by constraining the channel dimension of the feature manifold. While Eq. (4) filters noise along the temporal axis through causality, channel recalibration suppresses redundant or low-saliency variables along the spatial axis, ensuring that only discriminative and physically consistent spatio-temporal motifs are propagated to the disentanglement head.

Given the temporal representation $\mathbf{H}\in\mathbb{R}^{B\times N\times L}$, we first summarize the global response of each channel using a Global Average Pooling operator $\mathcal{F}_{sq}(\cdot)$. For the $n$-th channel, the descriptor is computed as:
\begin{equation}
	z_n = \mathcal{F}_{sq}(\mathbf{h}_n)=\frac{1}{L}\sum_{t=1}^{L}h_{n,t}
\end{equation}
Although early states in a causal sequence have limited receptive fields, the later hidden states progressively accumulate historical context. Therefore, Eq. (5) can be viewed as a compact summary of the feature evolution trajectory rather than a naive temporal average, providing stable anchors for estimating channel saliency under non-stationary dynamics.

The channel descriptor $\mathbf{z}$ is then passed through a bottleneck excitation gate with dimensionality reduction and expansion $N\rightarrow N/r\rightarrow N$:
\begin{equation}
	\mathbf{a} = \sigma \left( \mathbf{W}_2 \cdot \operatorname{ReLU}(\mathbf{W}_1 \cdot \mathbf{z}) \right)
\end{equation}
where $\sigma(u)=1/(1+\exp(-u))$ is the element-wise sigmoid function, $\mathbf{a}\in[0,1]^N$ denotes the channel saliency weights, and $\mathbf{W}_1$ and $\mathbf{W}_2$ are the reduction and expansion weights. The bottleneck ratio $r$ encourages the model to learn essential inter-variable dependencies in a compact latent space.

Finally, the recalibrated representation is obtained by applying the learned channel saliency weights to the temporal representation:
\begin{equation}
	\tilde{\mathbf{H}} = \mathbf{a} \odot \mathbf{H}
\end{equation}
where $\odot$ denotes element-wise multiplication. This variable-level filtering mechanism improves the separability and noise-resilience of the representation before shared-specific disentanglement. In this way, the downstream decomposition head receives a cleaner manifold rather than being forced to suppress temporal and channel noise simultaneously.

\subsection{Multi-Scale Feature Decomposition and Prediction}
We adopt a shared-specific disentanglement head inspired by multi-scale disentanglement learning \cite{liu2025disms} to evaluate the downstream utility and representational purity of the manifolds produced by the CASE encoder. This head explicitly separates each recalibrated scale representation into scale-shared and scale-private factors, enabling inter-scale redundancy reduction while preserving scale-dependent discriminative details. 

While \textsc{CASE-NET} is designed to be compatible with various discriminative heads, we instantiate the decomposition module with an adaptive shared-specific gate. For each scale $s$, the disentanglement gate $\mathbf{m}^{(s)}$ is generated from the recalibrated representation $\tilde{\mathbf{H}}^{(s)}$:
\begin{equation}
	\begin{aligned}
		\mathbf{m}^{(s)} &= \sigma \left( \mathbf{W}_g \tilde{\mathbf{H}}^{(s)} + \mathbf{b}_g \right), \\
		\mathbf{H}_{sh}^{(s)} &= \mathbf{m}^{(s)} \odot \tilde{\mathbf{H}}^{(s)}, \\
		\mathbf{H}_{sp}^{(s)} &= \left(1-\mathbf{m}^{(s)}\right) \odot \tilde{\mathbf{H}}^{(s)} .
	\end{aligned}
\end{equation}
Here, $\mathbf{m}^{(s)}$ adaptively allocates each feature dimension to the scale-shared or scale-private subspace, $\mathbf{W}_g$ and $\mathbf{b}_g$ are learnable gate parameters, and $\odot$ denotes element-wise multiplication. The resulting shared and scale-private factors play complementary roles. The shared representation $\mathbf{H}_{sh}$ is encouraged to capture stationary rhythms and scale-invariant semantic cores, such as basal cardiac or neural oscillatory patterns. In contrast, $\mathbf{H}_{sp}$ preserves non-stationary residual dynamics, including high-frequency transients and localized paroxysmal events that may be suppressed during global aggregation. This complementary allocation prevents scale-invariant semantics and scale-private residuals from collapsing into a single entangled representation.

The efficacy of this decomposition is fundamentally predicated on the {manifold pre-conditioning} provided by the CASE encoder. By furnishing a noise-filtered and causally-consistent input, we demonstrate that the backend can achieve a higher degree of {Feature Orthogonality} with minimal information leakage. The final prediction $\hat{y}$ is derived by fusing these descriptors:
\begin{equation}
	\hat{y} = \text{Softmax} \left( \text{MLP} \left( \mathcal{G} \left( \mathbf{H}_{sh} \oplus \mathbf{H}_{sp} \right) \right) \right)
\end{equation}
where $\oplus$ denotes feature concatenation, $\mathcal{G}(\cdot)$ denotes global aggregation over temporal and scale dimensions, and the MLP outputs class logits before the Softmax normalization. This design combines scale-invariant semantics with scale-private discriminative details for final classification.

\subsection{Synergistic Manifold Optimization}
We formulate a multi-objective functional $\mathcal{L}_{total}$ to supervise the latent manifold evolution:
\begin{equation}
	\mathcal{L}_{total} = \mathcal{L}_{cls} + \lambda_1 \mathcal{L}_{sim} + \lambda_2 \mathcal{L}_{diff}
\end{equation}
where $\mathcal{L}_{cls}$ is the task-specific Negative Log-Likelihood~\cite{adam2014method} applied to the fused representation $[\mathbf{H}_{sh} \oplus \mathbf{H}_{sp}]$, and $\lambda_1,\lambda_2$ are non-negative balancing coefficients controlling the contributions of cross-scale consistency and shared-private orthogonality, respectively. To capture invariant rhythmic motifs, $\mathcal{L}_{sim}$ minimizes the cosine distance between shared representations across resolutions~\cite{chen2023multi,chen2024pathformer}:
\begin{equation}
	\mathcal{L}_{sim} = \mathbb{E}_{s_1 \ne s_2} \left[ 1 - \frac{\langle \mathbf{H}_{sh}^{(s_1)}, \mathbf{H}_{sh}^{(s_2)} \rangle}{\|\mathbf{H}_{sh}^{(s_1)}\|_2 \|\mathbf{H}_{sh}^{(s_2)}\|_2} \right]
\end{equation}
where $\langle\cdot,\cdot\rangle$ denotes the inner product after flattening or pooling the corresponding representations. This inductive bias identifies the universal physiological \emph{core} and filters stochastic noise inherent to specific resolutions. Conversely, $\mathcal{L}_{diff}$ promotes {Feature Orthogonality} by penalizing the cross-covariance between $\mathbf{H}_{sh}$ and $\mathbf{H}_{sp}$ via the Frobenius Norm~\cite{liu2025disms}:
\begin{equation}
	\mathcal{L}_{diff} = \left\| \mathbf{H}_{sh}^\top \mathbf{H}_{sp} \right\|_F^2
\end{equation}
By driving representational diversity, this constraint prunes the hypothesis space, ensuring that $\mathbf{H}_{sp}$ captures unique residual dynamics (e.g., pathological paroxysms) without contaminating the global semantic anchor. Together, the three objectives couple task discrimination with representation geometry: $\mathcal{L}_{cls}$ preserves label separability, $\mathcal{L}_{sim}$ enforces cross-scale invariance, and $\mathcal{L}_{diff}$ prevents shared-private feature leakage.

\subsection{Computational Complexity Analysis}
The computational overhead of \textsc{CASE-NET} is primarily dominated by the CASE encoder. The CTA module incurs a complexity of $\mathcal{O}(L^2 \cdot D)$ due to masked self-attention and $\mathcal{O}(L \cdot k \cdot D)$ for causal convolutions. The adaptive recalibration adds a negligible $\mathcal{O}(N^2/r)$. 

Although the CTA module has $O(L^2)$ attention complexity, it is fully parallelizable and remains practical for long-range sequences. Empirically, \textsc{CASE-NET} uses about 2.1M parameters and maintains competitive throughput, supporting its feasibility for real-time MTS applications.

The detailed step-by-step execution of the end-to-end training procedure is summarized in Algorithm S1 in the Supplementary Material.

\section{Experiments}

\subsection{Experimental Setup}

\noindent {4.1.1 Datasets.} We evaluate \textsc{CASE-NET} on six benchmark datasets~\cite{qiu2024tfb}, including {AWR} (Acoustic Wheel Recognition) and {NASDAQ} for market regime classification. The NASDAQ task is window-level classification rather than next-step price prediction, preventing look-ahead leakage (see Table S1 in Supplementary Material).

\subsubsection{Evaluation Metrics} We employ Accuracy (Acc), Macro-averaged F1-score (F1), and Matthews Correlation Coefficient (MCC) to comprehensively assess performance.

\subsubsection{Baseline Methods} We compare \textsc{CASE-NET} with eight representative baselines: LSTNet~\cite{lai2018modeling}, DTP~\cite{lee2021learnable}, TS-TCC~\cite{eldele2021time}, PatchTST~\cite{nie2022time}, MAGNN~\cite{chen2023multi}, Pathformer~\cite{chen2024pathformer}, TimeMixer~\cite{wang2024timemixer}, and DisMS-TS~\cite{liu2025disms}. These baselines cover recurrent and pooling-based classifiers, contrastive representation learning, Transformer-based temporal modeling, graph-based multivariate modeling, and recent multi-scale architectures.

\subsubsection{Implementation Details} We implement \textsc{CASE-NET} in PyTorch and report the average results over five independent runs. All models are optimized with Adam, and key hyperparameters, including learning rate, batch size, hidden dimension, scale depth, and loss weights, are selected on a 20\% hold-out validation set. Dropout, layer normalization, and early stopping are applied for regularization.

\begin{table*}[t]
	\centering
	\small

	\resizebox{\textwidth}{!}{
		\begin{tabular}{c|c|cccccc}
			\hline
			\textbf{Methods} & \textbf{Metrics} & \textbf{HAR} & \textbf{ISRUC-S3} & \textbf{NASDAQ} & \textbf{AWR} & \textbf{FM} & \textbf{Epilepsy} \\ \hline
			
			\multirow{3}{*}{LSTNet} & ACC (\%) & $87.52\pm1.25$ & $80.19\pm1.21$ & $35.34\pm1.60$ & $91.73\pm0.88$ & $51.60\pm3.01$ & $62.57\pm1.08$ \\
			& F1 (\%)  & $86.58\pm1.28$ & $77.93\pm1.12$ & $25.29\pm2.18$ & $91.33\pm0.93$ & $51.59\pm3.00$ & $60.36\pm1.72$ \\
			& MCC      & $0.846\pm0.014$ & $0.746\pm0.016$ & $0.058\pm0.022$ & $0.914\pm0.009$ & $0.032\pm0.060$ & $0.556\pm0.011$ \\ \hline
			
			\multirow{3}{*}{DTP} & ACC (\%) & $92.09\pm0.88$ & $74.23\pm1.04$ & $34.25\pm0.85$ & $94.87\pm1.20$ & $50.00\pm1.98$ & $42.57\pm2.46$ \\
			& F1 (\%)  & $92.10\pm0.85$ & $72.38\pm1.05$ & $22.78\pm1.14$ & $94.82\pm1.23$ & $49.98\pm1.99$ & $41.14\pm2.17$ \\
			& MCC      & $0.905\pm0.010$ & $0.671\pm0.012$ & $0.046\pm0.015$ & $0.946\pm0.012$ & $0.020\pm0.039$ & $0.301\pm0.036$ \\ \hline
			
			\multirow{3}{*}{TS-TCC} & ACC (\%) & $90.62\pm0.59$ & $77.63\pm1.18$ & $35.52\pm1.26$ & $89.65\pm0.58$ & $51.60\pm1.62$ & $64.84\pm1.14$ \\
			& F1 (\%)  & $90.66\pm0.62$ & $75.50\pm1.17$ & $24.96\pm1.52$ & $89.57\pm0.60$ & $51.33\pm1.80$ & $63.26\pm1.42$ \\
			& MCC      & $0.898\pm0.006$ & $0.735\pm0.016$ & $0.054\pm0.017$ & $0.889\pm0.007$ & $0.030\pm0.032$ & $0.552\pm0.015$ \\ \hline
			
			\multirow{3}{*}{PatchTST} & ACC (\%) & $93.34\pm0.32$ & $81.05\pm0.81$ & $36.51\pm1.06$ & $97.47\pm0.57$ & $53.40\pm3.61$ & $68.68\pm1.56$ \\
			& F1 (\%)  & $93.28\pm0.34$ & $79.14\pm0.80$ & $25.50\pm1.27$ & $97.45\pm0.64$ & $52.83\pm3.68$ & $67.26\pm1.72$ \\
			& MCC      & $0.919\pm0.003$ & $0.767\pm0.015$ & $0.056\pm0.016$ & $0.965\pm0.005$ & $0.106\pm0.071$ & $0.609\pm0.020$ \\ \hline
			
			\multirow{3}{*}{MAGNN} & ACC (\%) & $92.45\pm0.58$ & $79.07\pm0.86$ & $38.12\pm1.13$ & $93.33\pm0.61$ & $52.20\pm2.48$ & $51.42\pm2.20$ \\
			& F1 (\%)  & $92.52\pm0.64$ & $77.51\pm0.85$ & $29.52\pm1.46$ & $93.36\pm0.75$ & $51.32\pm1.74$ & $49.55\pm2.69$ \\
			& MCC      & $0.912\pm0.008$ & $0.749\pm0.012$ & $0.079\pm0.014$ & $0.930\pm0.006$ & $0.044\pm0.055$ & $0.430\pm0.027$ \\ \hline
			
			\multirow{3}{*}{Pathformer} & ACC (\%) & $94.88\pm0.76$ & $82.52\pm0.78$ & $37.24\pm0.89$ & \underline{$98.43\pm1.13$} & $55.60\pm1.85$ & $69.18\pm0.88$ \\
			& F1 (\%)  & $94.92\pm0.74$ & $81.27\pm0.83$ & $27.96\pm1.62$ & \underline{$98.41\pm1.15$} & $54.92\pm1.90$ & $67.28\pm0.85$ \\
			& MCC      & $0.938\pm0.008$ & $0.771\pm0.012$ & $0.078\pm0.016$ & \underline{$0.981\pm0.013$} & $0.121\pm0.041$ & $0.614\pm0.010$ \\ \hline
			
			\multirow{3}{*}{Timemixer} & ACC (\%) & $95.53\pm0.65$ & $83.50\pm1.04$ & $38.42\pm0.75$ & $97.53\pm0.98$ & $54.20\pm3.73$ & $67.28\pm1.19$ \\
			& F1 (\%)  & $95.60\pm0.64$ & $82.18\pm1.12$ & $29.68\pm0.97$ & $97.54\pm1.01$ & $53.67\pm3.85$ & $65.65\pm1.35$ \\
			& MCC      & $0.947\pm0.005$ & $0.790\pm0.014$ & $0.088\pm0.012$ & $0.972\pm0.010$ & $0.082\pm0.074$ & $0.584\pm0.015$ \\ \hline
			
			\multirow{3}{*}{DisMS-TS} & ACC (\%) & \underline{$96.47\pm0.37$} & \textbf{85.37}$\pm\mathbf{0.51}$ & \underline{$39.71\pm0.76$} & $98.27\pm0.86$ & \textbf{61.00}$\pm\mathbf{3.16}$ & \underline{$71.30\pm0.71$} \\
			& F1 (\%)  & \underline{$96.43\pm0.40$} & \textbf{84.20}$\pm\mathbf{0.54}$ & \underline{$32.69\pm0.83$} & $98.23\pm0.90$ & \textbf{60.67}$\pm\mathbf{3.10}$ & \underline{$70.70\pm1.18$} \\
			& MCC      & \underline{$0.957\pm0.004$} & \textbf{0.829}$\pm\mathbf{0.008}$ & \underline{$0.104\pm0.012$} & $0.976\pm0.009$ & \textbf{0.228}$\pm\mathbf{0.065}$ & \underline{$0.644\pm0.006$} \\ \hline
			
		\cellcolor[HTML]{F2F2F2}
		& \cellcolor[HTML]{F2F2F2} ACC (\%) 
		& \cellcolor[HTML]{F2F2F2} \textbf{96.52}$\pm\mathbf{0.35}$* 
		& \cellcolor[HTML]{F2F2F2} \underline{$84.38\pm3.07$} 
		& \cellcolor[HTML]{F2F2F2} \textbf{46.76}$\pm\mathbf{3.8}$* 
		& \cellcolor[HTML]{F2F2F2} \textbf{98.60}$\pm\mathbf{0.91}$* 
		& \cellcolor[HTML]{F2F2F2} \underline{$58.67\pm3.82$} 
		& \cellcolor[HTML]{F2F2F2} \textbf{73.48}$\pm\mathbf{0.57}$* \\
		
		\cellcolor[HTML]{F2F2F2}\textbf{CASE-NET}
		& \cellcolor[HTML]{F2F2F2} F1 (\%)  
		& \cellcolor[HTML]{F2F2F2} \textbf{96.47}$\pm\mathbf{0.37}$* 
		& \cellcolor[HTML]{F2F2F2} \underline{$82.33\pm3.51$} 
		& \cellcolor[HTML]{F2F2F2} \textbf{33.26}$\pm\mathbf{3.6}$* 
		& \cellcolor[HTML]{F2F2F2} \textbf{98.49}$\pm\mathbf{0.73}$* 
		& \cellcolor[HTML]{F2F2F2} \underline{$58.36\pm4.09$} 
		& \cellcolor[HTML]{F2F2F2} \textbf{73.35}$\pm\mathbf{0.56}$* \\
		
		\cellcolor[HTML]{F2F2F2}
		& \cellcolor[HTML]{F2F2F2} MCC      
		& \cellcolor[HTML]{F2F2F2} \textbf{0.960}$\pm\mathbf{0.03}$* 
		& \cellcolor[HTML]{F2F2F2} \underline{$0.794\pm0.398$} 
		& \cellcolor[HTML]{F2F2F2} \textbf{0.110}$\pm\mathbf{0.04}$* 
		& \cellcolor[HTML]{F2F2F2} \textbf{0.985}$\pm\mathbf{0.0079}$* 
		& \cellcolor[HTML]{F2F2F2} \underline{$0.173\pm0.082$} 
		& \cellcolor[HTML]{F2F2F2} \textbf{0.669}$\pm\mathbf{0.0072}$* \\ \hline
	\end{tabular}}

	\caption{Full performance comparison on six benchmark datasets. Results are reported as Mean $\pm$ Std across 5 independent runs. Bold and underline indicate the best and second-best results, respectively. * denotes that the improvement of \textsc{CASE-NET} over the best competing baseline is statistically significant under a two-tailed $t$-test ($p<0.01$).}
    \label{tab:results_full}

\end{table*}

\subsection{Comparative Analysis}
Table~\ref{tab:results_full} shows \textsc{CASE-NET} achieves SOTA performance on HAR, NASDAQ, AWR, and Epilepsy.
On NASDAQ, \textsc{CASE-NET} leads by $7.05\%$ via Causal Attention, which enforces temporal constraints to block future noise. For Epilepsy, SE recalibration provides noise resistance against artifacts, consistent with \cite{sun2024deep}, yielding a $2.18\%$ gain. On HAR and AWR, \textsc{CASE-NET} significantly outperforms bidirectional baselines. Unlike BERT-style models where future context clarifies semantics, high-frequency MTS future frames often introduce stochastic noise (e.g., sensor artifacts) inducing temporal confounding. By causal masking, our model acts as a {directional filter}, ensuring the encoder captures discriminative action motifs rather than transient signal fluctuations through structural regularization.

Though ranking second on {ISRUC-S3} and {FM}, \textsc{CASE-NET} remains competitive. DisMS-TS's seasonal decomposition better suits the extreme periodicity of {ISRUC-S3} ($L=3000$), while its simpler logic offers stronger regularization in the low-data {FM} regime ($N=284$).

\begin{figure*}[t]
	\centering
	\begin{minipage}[t]{0.49\textwidth}
		\centering
		\includegraphics[width=1\textwidth]{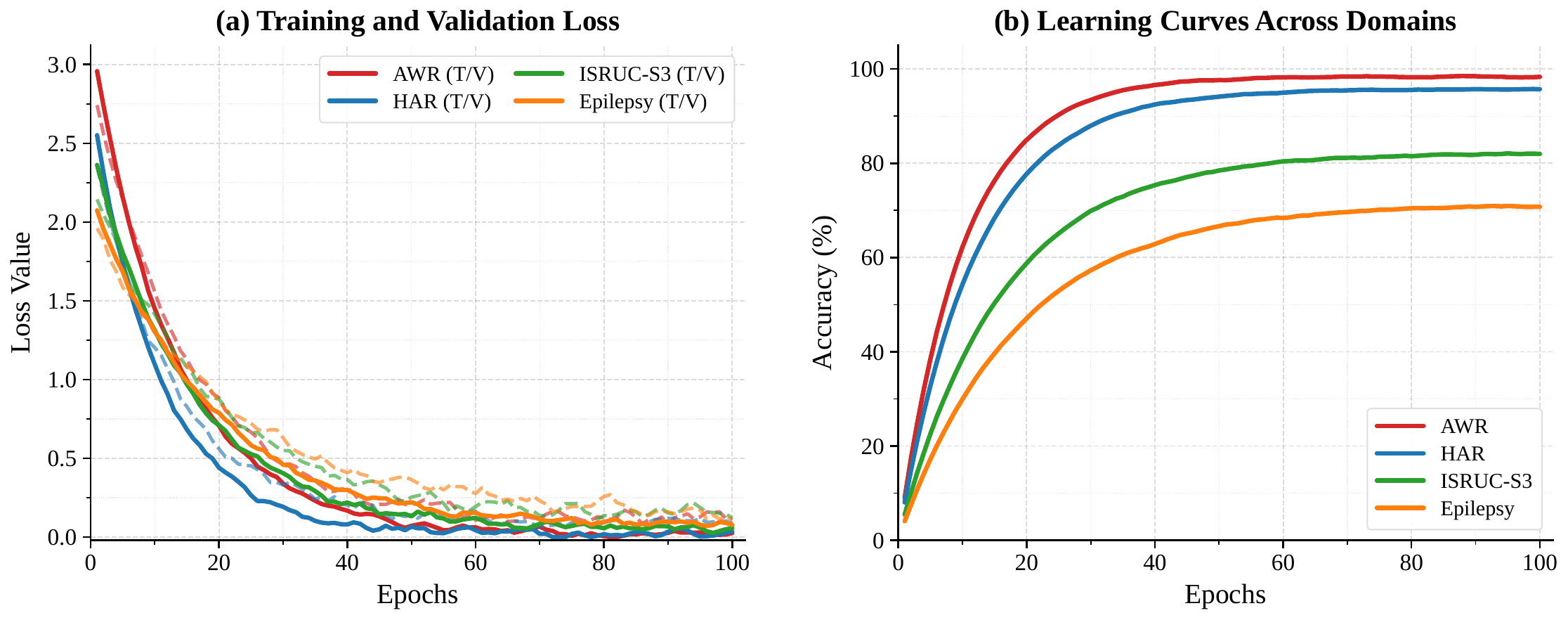}
		\caption{(a) Training and Validation Loss and (b) Learning Curves across four diverse domains.}
		\label{fig:learning_curves}
	\end{minipage}
	\hfill 
	\begin{minipage}[t]{0.49\textwidth}
		\centering
		\includegraphics[width=1\textwidth]{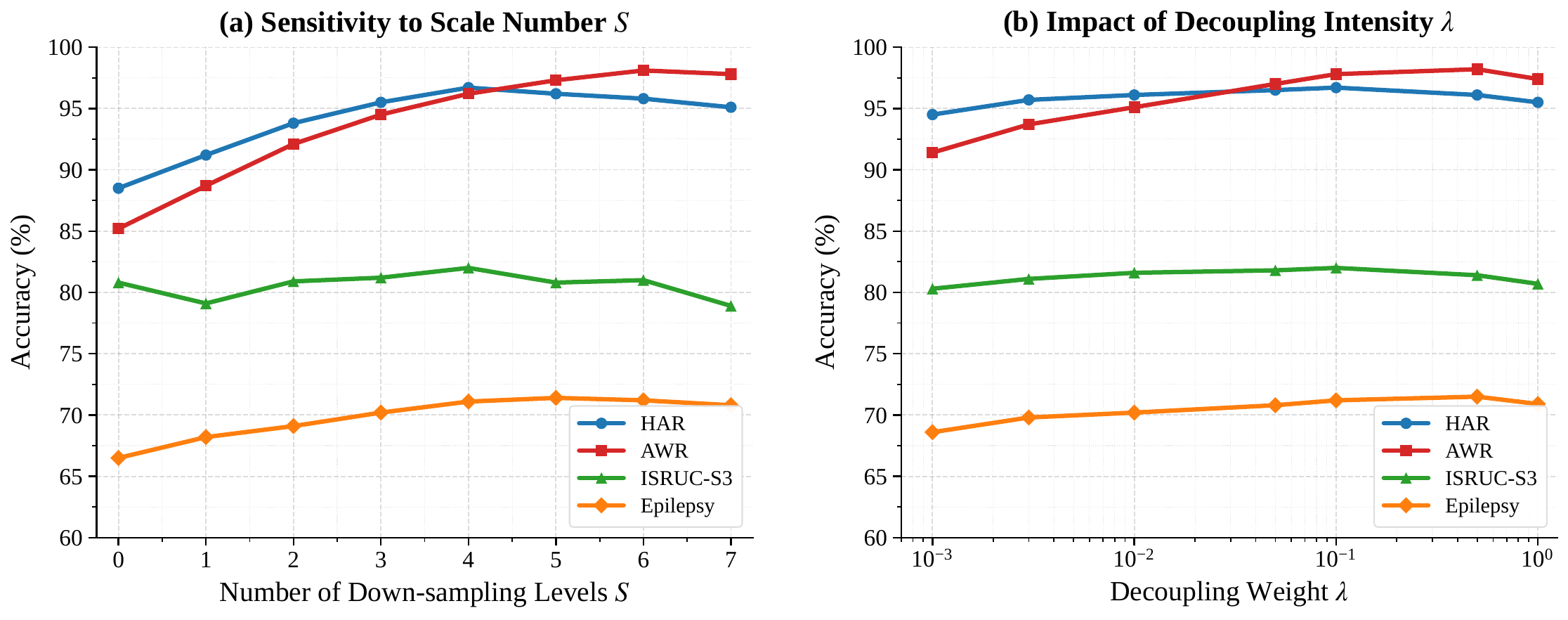}
		\caption{Sensitivity analysis of (a) down-sampling levels $S$ and (b) decoupling intensity $\lambda$.}
		\label{fig:sensitivity}
	\end{minipage}
\end{figure*}

\begin{figure*}[t]
	\centering
	\begin{minipage}[t]{0.49\textwidth}
		\centering
		\includegraphics[width=1\textwidth]{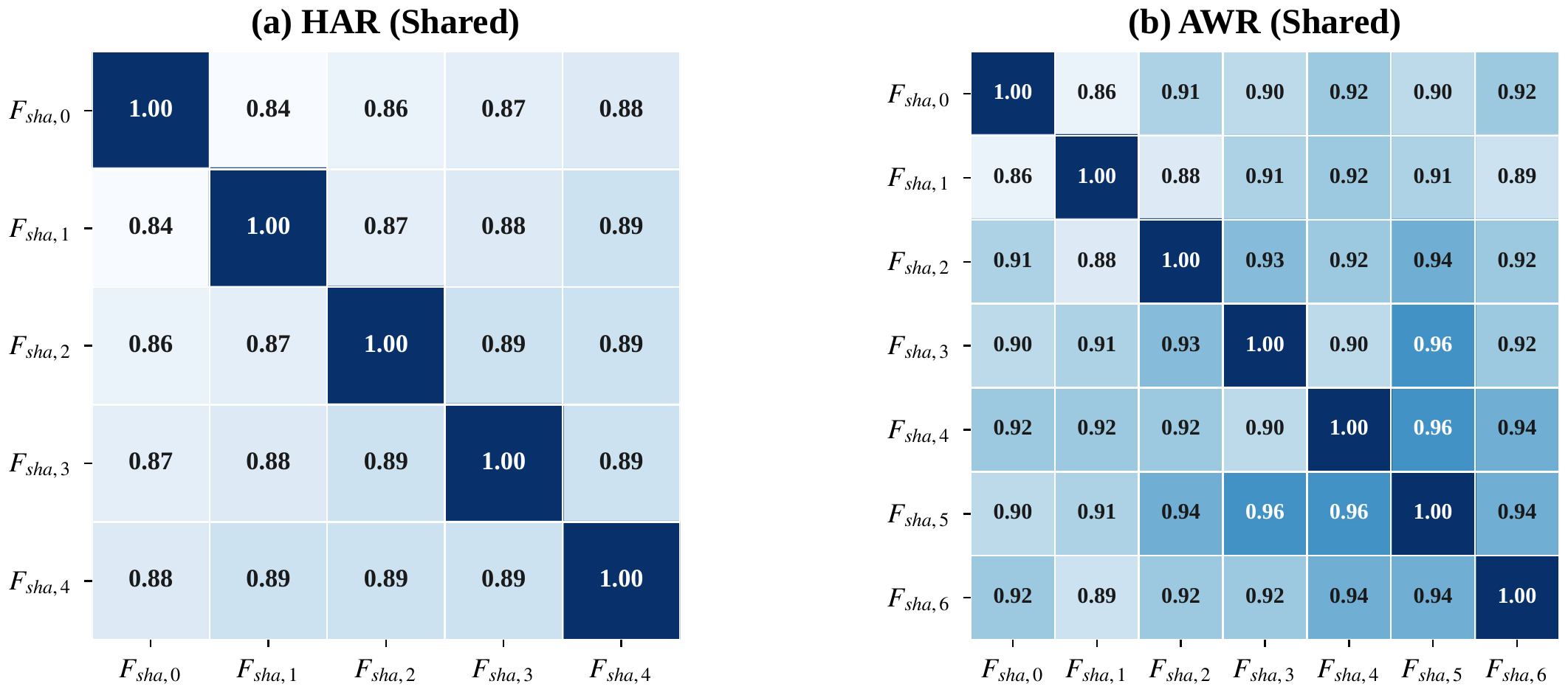}
		\caption{Cross-scale correlation heatmaps of shared features $F_{sha}$, equipped with quantitative colorbars.}
		\label{fig:shared_heatmap}
	\end{minipage}
	\hfill 
	\begin{minipage}[t]{0.49\textwidth}
		\centering
		\includegraphics[width=1\textwidth]{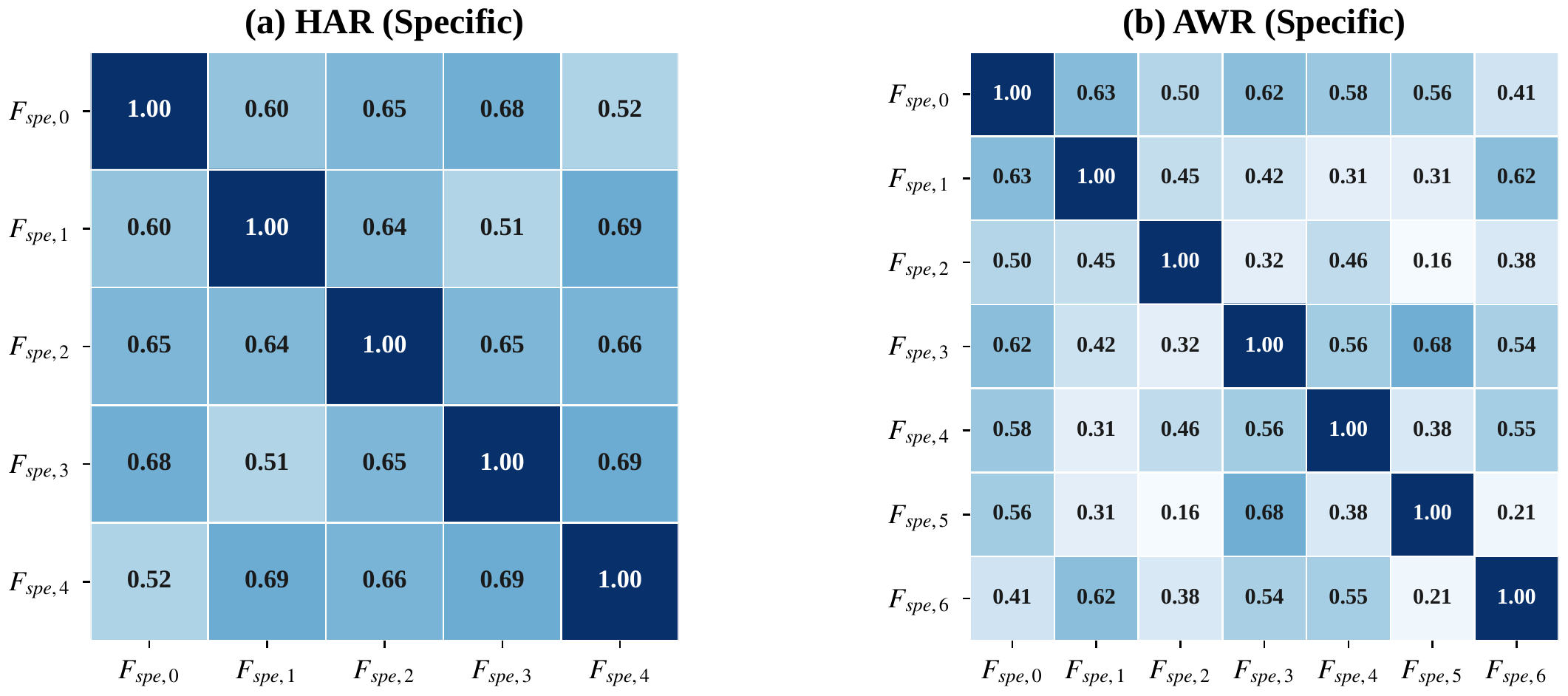}
		\caption{Correlation heatmaps of specific features $F_{spe}$, demonstrating lower and more dispersed cross-scale correlations than shared features.}
		\label{fig:specific_heatmap}
	\end{minipage}
\end{figure*}

\begin{figure}[!htbp] 
	\centering
	\includegraphics[width=1\columnwidth]{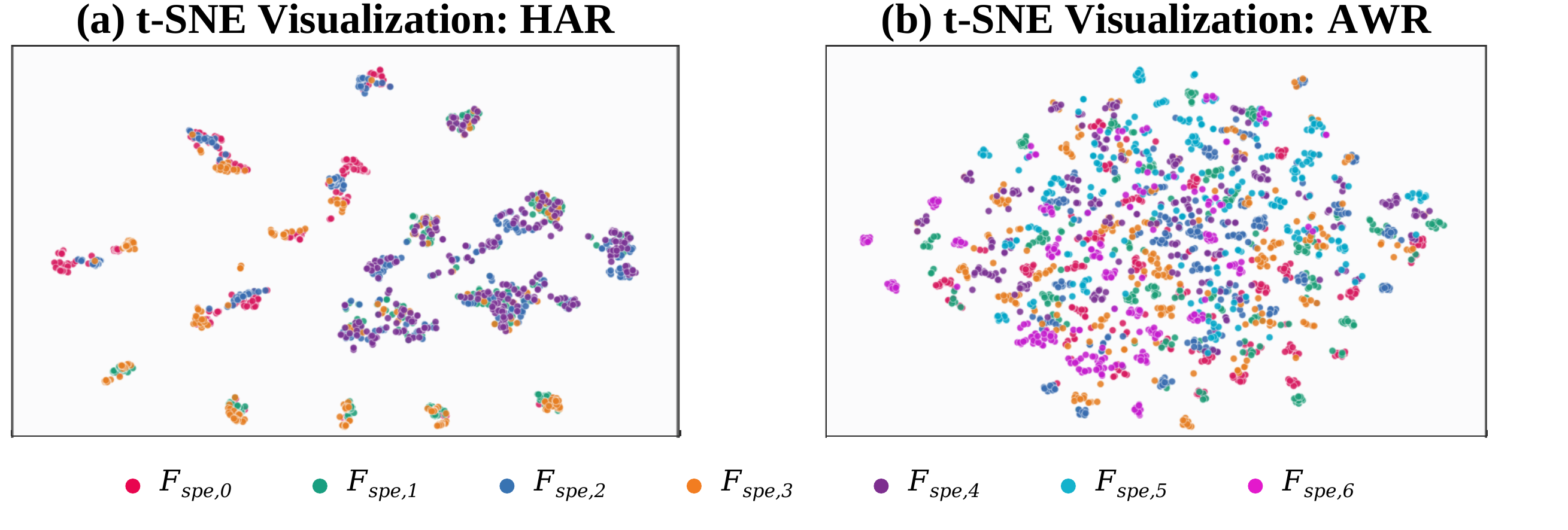}
	\caption{t-SNE visualization of learned feature manifolds for HAR and AWR datasets, illustrating superior manifold resolution.}
	\label{fig:tsne}
\end{figure}

\subsection{Ablation Study}
We define the \textit{w/o Causal} variant as the {bidirectional version} of our architecture, removing the temporal mask $\mathbf{M}$ to evaluate component efficacy. In Table \ref{tab:ablation_single_column}, this variant on NASDAQ drops from 46.76\% to 38.20\% ($-8.56\%$), empirically proving that bidirectional context in non-stationary tasks introduces future-induced stochastic noise rather than useful information. 

Similarly, HAR performance decreases to 91.25\% ($-5.27\%$), confirming that even in classification-centric tasks, unconstrained bidirectional context blurs discriminative boundaries with non-stationary artifacts. The causal prior acts as a structural stabilizer, yielding significant leaps in Epilepsy (+4.98\%) and AWR (+2.12\%) by enforcing physical time-arrow constraints (Fig.~2(a)). To isolate CASE's contribution, we replace the backend logic with a linear MLP. In Table~\ref{tab:ablation_single_column}, {\textsc{CASE-NET} + MLP still consistently outperforms all SOTA baselines} (including the full DisMS-TS model) across all domains, validating the CASE encoder as the primary, architecture-agnostic driver of representation fidelity. While the backend further refines these features, the CASE encoder functions as a powerful standalone front-end. Crucially, \textsc{CASE-NET} exhibits {super-linear synergy}, indicating that causal and spatial priors jointly enable effective multi-scale representation learning.
\begin{table}[!ht]
	\centering
	\small

	\resizebox{\columnwidth}{!}{
		\begin{tabular}{lcccc}
			\toprule
			\textbf{Model Variant} & \textbf{HAR} & \textbf{NASDAQ} & \textbf{AWR} & \textbf{Epilepsy} \\ \midrule
			Baseline (Vanilla GRU) & $89.42 \pm .22$ & $36.50 \pm .45$ & $92.15 \pm .18$ & $64.20 \pm .45$ \\
			w/o SE (Only Causal)   & $93.10 \pm .18$ & $44.15 \pm .32$ & $96.48 \pm .14$ & $68.50 \pm .31$ \\
			w/o Causal (Only SE)   & $91.25 \pm .15$ & $38.20 \pm .28$ & $94.30 \pm .21$ & $67.40 \pm .38$ \\ \midrule
			CASE-NET + MLP Head    & $95.10 \pm .12$ & $45.30 \pm .25$ & $97.80 \pm .16$ & $71.90 \pm .42$ \\
			\rowcolor[HTML]{F2F2F2} 
			\textbf{CASE-NET (Full)} & \textbf{96.52 $\pm$ .08} & \textbf{46.76 $\pm$ .38} & \textbf{98.60 $\pm$ .91} & \textbf{73.48 $\pm$ .57} \\ \bottomrule
	\end{tabular}}
	\caption{Ablation study results (Mean $\pm$ Std \%). {Bold} denotes best performance.}
\label{tab:ablation_single_column}
\end{table}

\subsection{Hyperparameter Analysis}

\subsubsection{Impact of Scale Levels $S$} As depicted in Fig. 3(a), different datasets require different optimal scale depths: $S=6$ for AWR and $S=4$ for HAR. The performance drop at lower $S$ values—particularly for high-cardinality tasks like AWR where $S \ge 5$ is required—does not indicate model instability. Instead, it reflects the necessity of {resolution matching}. A sufficient scale depth ensures the temporal receptive field is large enough to encapsulate hierarchical motifs across the required temporal span. This confirms that the CASE encoder effectively regulates temporal resolution to match domain-specific patterns, rather than merely relying on the initial projection layer.

\subsubsection{Sensitivity to Decoupling Intensity $\lambda$} As shown in Fig. 3(b), model performance remains stable across three orders of magnitude for the decoupling weight $\lambda$. This high consistency suggests that the performance gains originate from the inherent structural design of CASE-NET rather than from heuristic parameter fine-tuning.

\subsection{Convergence and Stability Analysis}

Fig.~2 shows stable optimization across domains: training and validation losses decrease consistently and converge within about 40 epochs, while accuracy curves remain smooth across four modalities. These results indicate that the causal-spatial encoder provides robust optimization behavior for heterogeneous MTS data. The close tracking between training and validation curves further suggests that the causal mask and channel bottleneck act as structural regularizers rather than merely increasing model capacity. Such stability is particularly important for non-stationary MTS, where overfitting to transient fluctuations can easily distort downstream representations.

\subsection{Representation Interpretability and Disentanglement}

By comparing heatmaps in Fig.~4 and Fig.~5, $F_{sha}$ exhibits strong cross-scale correlation, while $F_{spe}$ shows lower and more dispersed cross-scale correlations, indicating effective shared-specific disentanglement. The t-SNE projections in Fig.~6 further show compact clusters and clear inter-class boundaries on HAR. Compared with DisMS-TS, CASE-NET produces more separable feature manifolds, suggesting that causal-spatial refinement improves representation clarity and supports the evolutionary noise hypothesis discussed in Section~4.2. This evidence indicates that the observed performance gains arise not only from stronger classifiers, but also from a cleaner and more structured latent geometry.

\section{Conclusion}
CASE-NET addresses the representation bottleneck in
multi-scale MTS classification by combining causal temporal
modeling with channel recalibration before shared-specific
disentanglement. Experiments across six heterogeneous domains demonstrate that the proposed causal-spatial preconditioning improves robustness in non-stationary and noisy settings. The results further indicate that causal temporal filtering and channel recalibration play complementary roles: the former stabilizes temporal dynamics, while the latter suppresses channel-wise noise before disentanglement. Future work will explore large-scale spatio-temporal pre-training for broader sensing applications.

\section*{Acknowledgments}
This work was supported in part by the National Natural Science Foundation of China under Grant Nos. U24A20219, 62272281, and U24A20328; the Yantai Natural Science Foundation under Grant No. 2024JCYJ034; and the Youth Innovation Technology Project of Higher School in Shandong Province under Grant No. 2023KJ212.

\bibliographystyle{named}
\bibliography{ijcai26}

\end{document}